\documentclass[10pt,twocolumn,letterpaper]{article}

\usepackage{cvpr}              

\usepackage{csquotes}

\usepackage{epsfig}
\usepackage{graphicx}
\usepackage{amssymb}
\usepackage{enumitem}
\usepackage{bm}
\usepackage{subcaption}
\usepackage{multirow}
\usepackage{nicefrac}
\usepackage{booktabs}

\usepackage{xcolor}
\usepackage{listings}
\usepackage{xparse}
\NewDocumentCommand{\codeword}{v}{%
  \colorbox{lightgray}{\texttt{#1}}%
}

\definecolor{mygreen}{RGB}{51, 204, 51}
\definecolor{mypurple}{RGB}{204, 51, 204}

\definecolor{cvprblue}{rgb}{0.21,0.49,0.74}
\usepackage[pagebackref,breaklinks,colorlinks,allcolors=cvprblue]{hyperref}

\def\paperID{*****} 
\def\confName{CVPR}
\def\confYear{2025}

\title{Learning Camera Movement Control from Real-World Drone Videos}

\author{Yunzhong Hou$^1$ \quad Liang Zheng$^1$ \quad Philip Torr$^2$\\
$^1$Australian National University  \quad $^2$University of Oxford
}

\begin{document}

\twocolumn[{
\maketitle

\begin{center}
    \vskip -0.2in
    \centering
    \includegraphics[width=\linewidth]{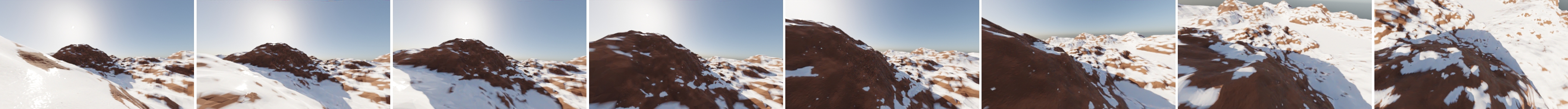}
    \includegraphics[width=\linewidth]{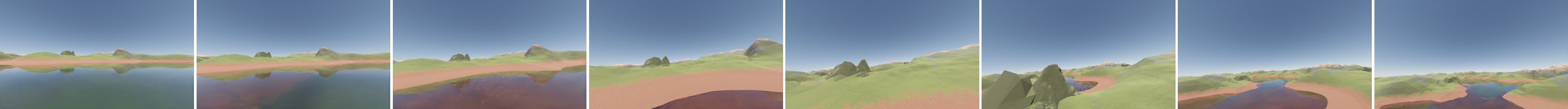}
    \includegraphics[width=\linewidth]{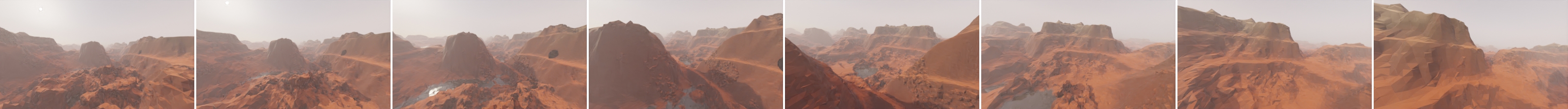}
    \includegraphics[width=\linewidth]{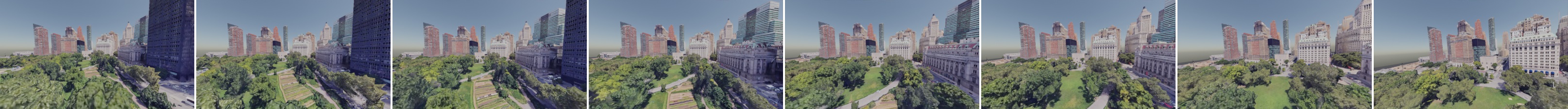}
    \includegraphics[width=\linewidth]{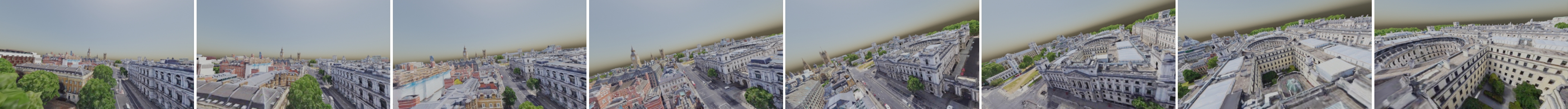}
    \includegraphics[width=\linewidth]{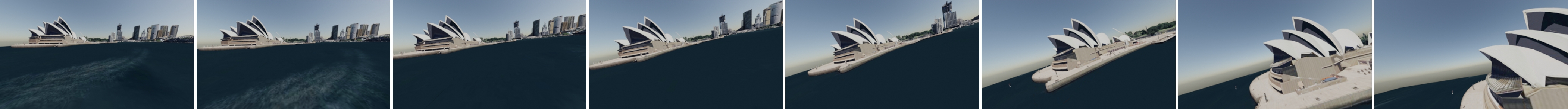}
   \vskip -0.05in
    \captionof{figure}{Examples of recorded videos from our AI cameraman. Instead of generating \textit{non-existent} content directly in the pixel space, our system outputs camera movements to film {\textit{existing} subjects into aesthetically pleasing videos}. 
    }
    \label{fig:banner}
\end{center}
}]

\vskip 0.1in
\begin{abstract}
   \vskip -0.15in

This study seeks to automate camera movement control for filming existing subjects into attractive videos, contrasting with the creation of non-existent content by directly generating the pixels. We select drone videos as our test case due to their rich and challenging motion patterns, distinctive viewing angles, and precise controls. Existing AI videography methods struggle with limited appearance diversity in simulation training, high costs of recording expert operations, and difficulties in designing heuristic-based goals to cover all scenarios. To avoid these issues, we propose a scalable method that involves collecting real-world training data to improve diversity, extracting camera trajectories automatically to minimize annotation costs, and training an effective architecture that does not rely on heuristics. Specifically, we collect 99k high-quality trajectories by running 3D reconstruction on online videos, connecting camera poses from consecutive frames to formulate 3D camera paths, and using Kalman filter to identify and remove low-quality data. Moreover, we introduce DVGFormer, an auto-regressive transformer that leverages the camera path and images from all past frames to predict camera movement in the next frame. We evaluate our system across 38 synthetic natural scenes and 7 real city 3D scans. We show that our system effectively learns to perform challenging camera movements such as navigating through obstacles, maintaining low altitude to increase perceived speed, and orbiting tower and buildings, which are very useful for recording high-quality videos. Data and code are available at \url{dvgformer.github.io}.
\end{abstract}

\vskip -0.15in
\section{Introduction}
\label{sec:intro}

Videography \cite{cubitt1993videography} captures \textit{existing} subjects in a visually attractive manner, \eg, moments in people's lives, places they have been, or things they have seen. While recent studies on AI generated content gain wide attention, most of them \cite{ho2020denoising, openaisora, polyak2024movie} create visual content that \textit{does not exist} by directly generating the RGB pixels. 
In this work, we aim to build an AI cameraman to capture \textit{existing} subjects. Specifically, it predicts camera movement in videos, a key aspect of videography that affects how audience experience content through the change in perspective \cite{acmiOnlineLearning,studiobinderWATCHUltimate}. We focus on drone videos, which not only contains rich and challenging camera trajectories including stimulating first-person-view (FPV) shots with drastic perspective changes, but also features distinctive viewing angles with airborne footage and precise controls over the camera location and direction. 

However, existing studies on AI videography or cinematography face a few important challenges. First, they are built on either simulation environments with game engines \cite{shah2018airsim,jeon2020detection,yu2022enabling} or 3D animations that contain recordings of human experts' camera operations \cite{jiang2020example}. Such data either offer limited appearance changes or are very costly to collect. Another solution is to collect real-world data with teleoperations from human experts like in robotics \cite{rt12022arxiv,walke2023bridgedata}, but is also expensive. 
Second, current studies on AI videography rely heavily on human heuristics, \eg, controlling the camera angle and distance to keep the actor within frame \cite{joubert2016towards, nageli2017real, huang2018act}. Despite the best effort, it remains difficult to exhaust every possible scenario or to capture the finer details with heuristic-based goals. 

Aiming at building an effective AI cameraman, we 
investigate scalable approaches to both training data and architecture. 
On the one hand, we collect the DroneMotion-99k dataset, which has 99,003 camera trajectories from real-world videos with a total duration of over 180 hours.   
For trajectory annotation, instead of recording expensive human expert operations, we propose an economical way to extract ground-truth camera operations from online videos with minimal human annotation. 
Specifically, we split the scraped YouTube videos into clips and run Colmap \cite{schoenberger2016sfm} reconstruction to recover the 3D camera pose for each frame. We then build the camera path by connecting the camera poses from consecutive frames and apply Kalman filters \cite{wan2000unscented} to identify and discard low-quality reconstructions. 

On the other hand, we introduce Drone VideoGraphy transFormer, or DVGFormer, an auto-regressive transformer \cite{radford2019language,chatgpt,touvron2023llama} for camera path prediction. Based on inputs of camera poses, motions, image patches, and depth estimations from previous frames, DVGFormer outputs camera motion for the next frame. While existing works on generalize robotics models can only process one or a few frames and only rely on images to describe the past \cite{rt12022arxiv,rt22023arxiv,kim2024openvla,shah2023vint}, our method takes a 10-second video as input and processes both images and camera path, so that long-term dependencies are considered and trajectory smoothness is improved.

We train DVGFormer using the DroneMotion-99k dataset. During inference, given 3D assets and initial camera pose, we predict a camera trajectory by iteratively predicting camera motion for the next frame. This trajectory can be conveniently rendered into a high-quality video without post-processing or modification in the pixel space. Sample video frames are shown in Fig.~\ref{fig:banner}. Compared with a baseline method inspired by \cite{rt22023arxiv}, the proposed system has significantly better user preference, lower collision rate and higher motion smoothness.  

\section{Related Work}

\textbf{Video generation and camera movement conditioning} 
has become a very hot topic, and most of these works focus on generating content that \textit{does not exist} \cite{blattmann2023svd, guo2023animatediff, openaisora, polyak2024movie}. 
InfiniteNature-Zero \cite{li2022infinitenature} sample camera trajectories and learn to render the perspective view. 
Valevski \etal \cite{valevski2024diffusionmodelsrealtimegame} jointly train an agent to play video games and collect training data, and a diffusion model to predict the next frame. 
Many recent study investigate how to take user-specified camera path as condition for video generation \cite{zhang2024recapture,namekata2024sg,kuang2024collaborative,he2024cameractrl}. 
It should be noted that they require \textit{predefined} camera paths, while our goal is to \textit{generate} the 3D camera trajectory for capturing good videos. 

\begin{figure*}
    \centering
    \includegraphics[width=\linewidth]{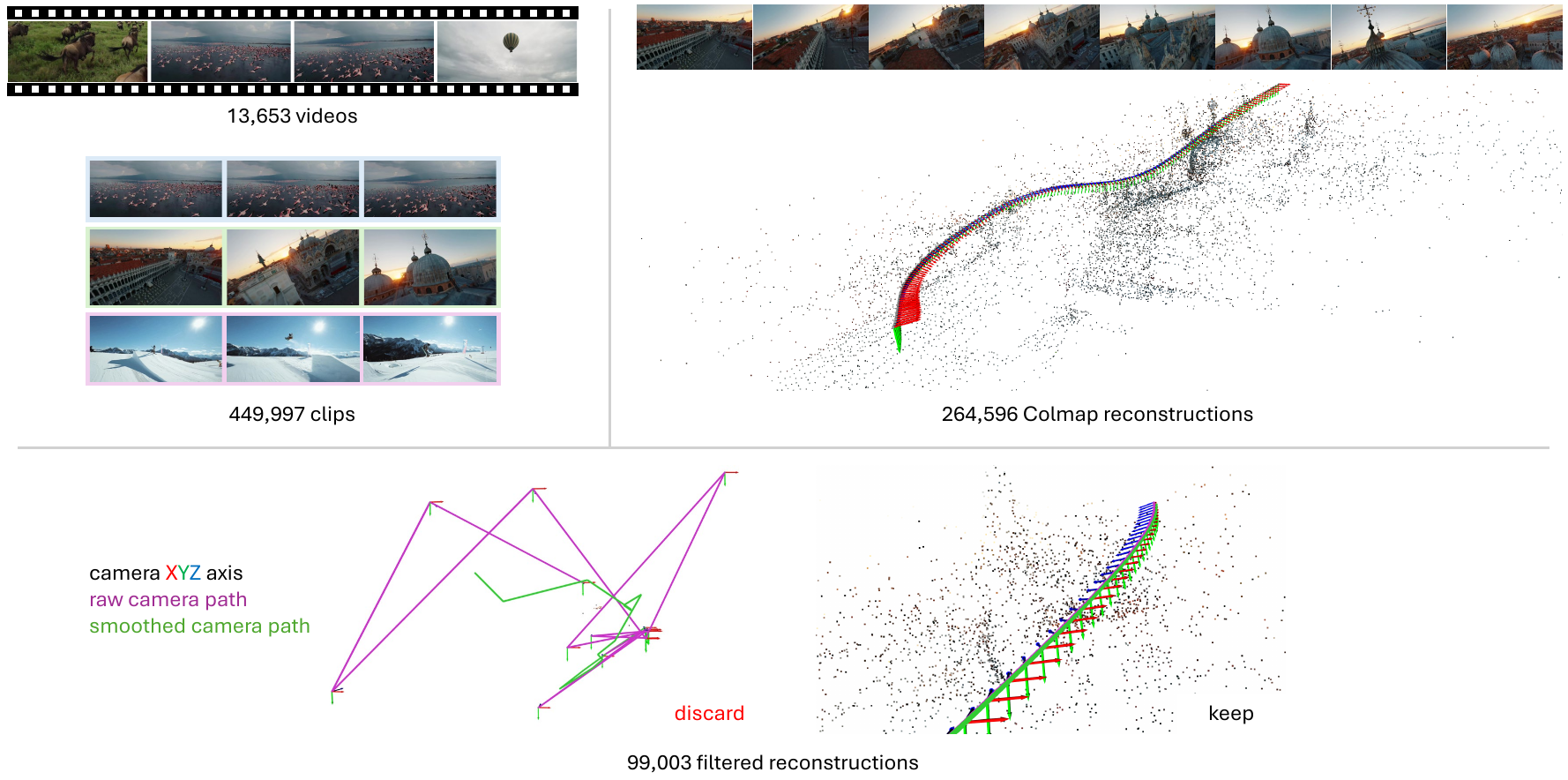}
    \caption{Data collection pipeline. \textbf{Top left}: For scraped YouTube videos, we  run shot change detection \cite{PySceneDetect} to split the videos into clips of individual scene. \textbf{Top right}: We then use Colmap \cite{schoenberger2016sfm} to reconstruct the 3D scene and recover camera poses from video frames. \textbf{Bottom}: Finally, we connect camera poses from consecutive frames to formulate 3D camera  trajectories and apply Kalman filter \cite{wan2000unscented} to discard low quality reconstructions whose camera poses from neighboring frames are drastically different. }
    \label{fig:data}
\end{figure*}

\textbf{AI cinematography and videography} record attractive videos from \textit{the existing}, \eg, sports, events, landscapes, or vlogs. 
Most works are built with human heuristics, such as keeping the actors in frame and maintaining distance  \cite{10.1145/237170.237259,yu2022bridging,yu2022enabling,azzarelli2024reviewing}.
Instead of heuristics, Jiang \etal \cite{jiang2020example} directly learn the camera controls from animation scenes created by 3D artists. 
For drone videos, most still follow the heuristic based approach \cite{joubert2016towards, nageli2017real, huang2018act, mademlis2019high, mademlis2024vision} and use simulation training from the AirSim \cite{shah2018airsim}  platform  \cite{jeon2020detection, jeon2020integrated,pueyo2024cinempc}. Huang \etal \cite{huang2019learning2} learn the optical flow as surrogates of the camera pose. Ashtari \etal collect the DVCD18K dataset \cite{ashtari2022drone} from online videos and use SLAM \cite{mur2017orb} to recover camera pose, but without camera intrinsics, the SLAM reconstruction quality is poor. 
These approaches remain \textit{difficult to scale} because they either rely on simulated training data with limited appearance change, or 3D animations that are costly to collect, and are often built with handcrafted heuristics.


\textbf{Generalist robotics models} aim to solve tasks including picking, moving, and placing objects in different scenarios by \textit{scaling both data and model}. 
In terms of data, RT-1 collect 130k real-world demonstrations performed by human operators \cite{rt12022arxiv} and BridgeDataV2 \cite{walke2023bridgedata} open source 60k teleoperation trajectories. 
In terms of architecture, most either use a one \cite{rt22023arxiv,kim2024openvla} or a few past frames \cite{rt12022arxiv,shah2023vint} to describe the past information, and directly predict the robot action via behavior cloning. 
Several others \cite{bruce2024genie,ye2024latent} also attempts to learn latent actions that encode the transition between frames. 
While we aim to \textit{mimic the scalable approach} in recent studies, there are still several challenges. First, recording teleoperation from human experts is {expensive}. Second, the generalist robotics models mostly consider a {short horizon} and the past \textit{images only}, both of which might be less effective for videography. 

\section{The DroneMotion-99K Dataset}
\label{sec:data}
To build an AI cameraman in a scalable manner, our first step is to collect high-quality real-world training data. 
Specifically, we develop a pipeline that automatically generates 3D camera paths from online videos. Unlike other robotics tasks, \eg, picking, moving, and placing, that require a separate recording of teleoperations from human operators, for AI videography, the camera movement can be retrieved via 3D reconstruction and does not need any external recordings of the operations. Because of this automatic approach, we have a much wider range of data to choose from, \eg, YouTube videos, at a much cheaper cost. 
An overview of our data collection pipeline is shown in Fig.~\ref{fig:data}.


\subsection{Video Preprocessing}
\label{subsec:video preprocess}
We first build a drone footage database from YouTube videos.
We filter out the videos for harmful or weaponized usage of drones and download 13k appropriate videos at 1080p resolution for a total duration of 1.5k hours or 62 days. 
Once we have the videos, we then split each videos into clips of an individual scene or \textit{``shot''}, which refers to a continuous sequence of frames captured by a single camera without interruption \cite{bordwell2010film} (see Fig.~\ref{fig:data} top left). In fact, the average durations of our scraped YouTube videos are roughly 400 seconds, whereas the average shot lengths in films nowadays can be as short as several seconds \cite{cutting2011quicker}. After shot change detection with PySceneDetect \cite{PySceneDetect}, we end up with {\raise.17ex\hbox{$\scriptstyle\sim$}}450k video clips. We then filter out clips with conversations since they are often unrelated. 

\subsection{3D Reconstruction}
\label{subsec:3d recons}
Despite the popularity and efficiency of SLAM methods \cite{campos2021orb,teed2021droid} in robotics tasks, for 3D reconstruction of online videos, their performance is often limited due to the lack of camera intrinsics \cite{ashtari2022drone}. Thus, we use the Structure-from-Motion (SfM) method  Colmap \cite{schoenberger2016sfm} for this task (Fig.~\ref{fig:data} top right). We tune several settings to balance the computational cost and the reconstruction quality. For frame extraction, we choose an intermediate frame rate of 15 fps to reduce computation and maintain a high image resolution at 1080p to ensure the quality of image feature points. For SIFT \cite{lowe2004distinctive} features, we prioritize stronger features with affine-covariant feature detection and Domain-Size Pooling \cite{dong2015domain} over the number of feature points per image. Both of these options help the 3D reconstruction quality, but the latter can be more computationally expensive. Since our focus is not on the 3D point clouds, we limit the number of feature points to 512 per image. Using 4 threads per process, each Colmap reconstruction worker takes roughly 200 seconds to finish on average. In total, the computation takes {\raise.17ex\hbox{$\scriptstyle\sim$}}34k CPU hours or roughly three weeks on a 224-core CPU server, producing {\raise.17ex\hbox{$\scriptstyle\sim$}}250k reconstructions. 

Once we have the 3D reconstructions, we connect the camera pose of consecutive frames to form the camera trajectories. To solve the scale ambiguities, we normalize the reconstructed scene based on the average camera distance between neighboring frames, assuming that the drone moving speed is stable across different videos. 

\begin{figure}
    \centering
    \includegraphics[width=0.69\linewidth]{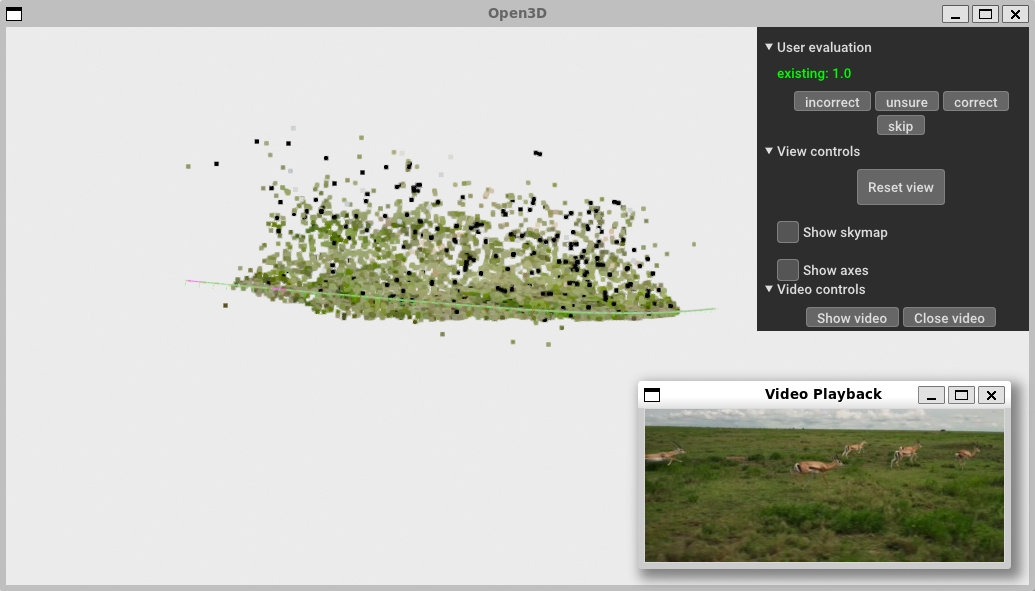}
    \includegraphics[width=0.29\linewidth]{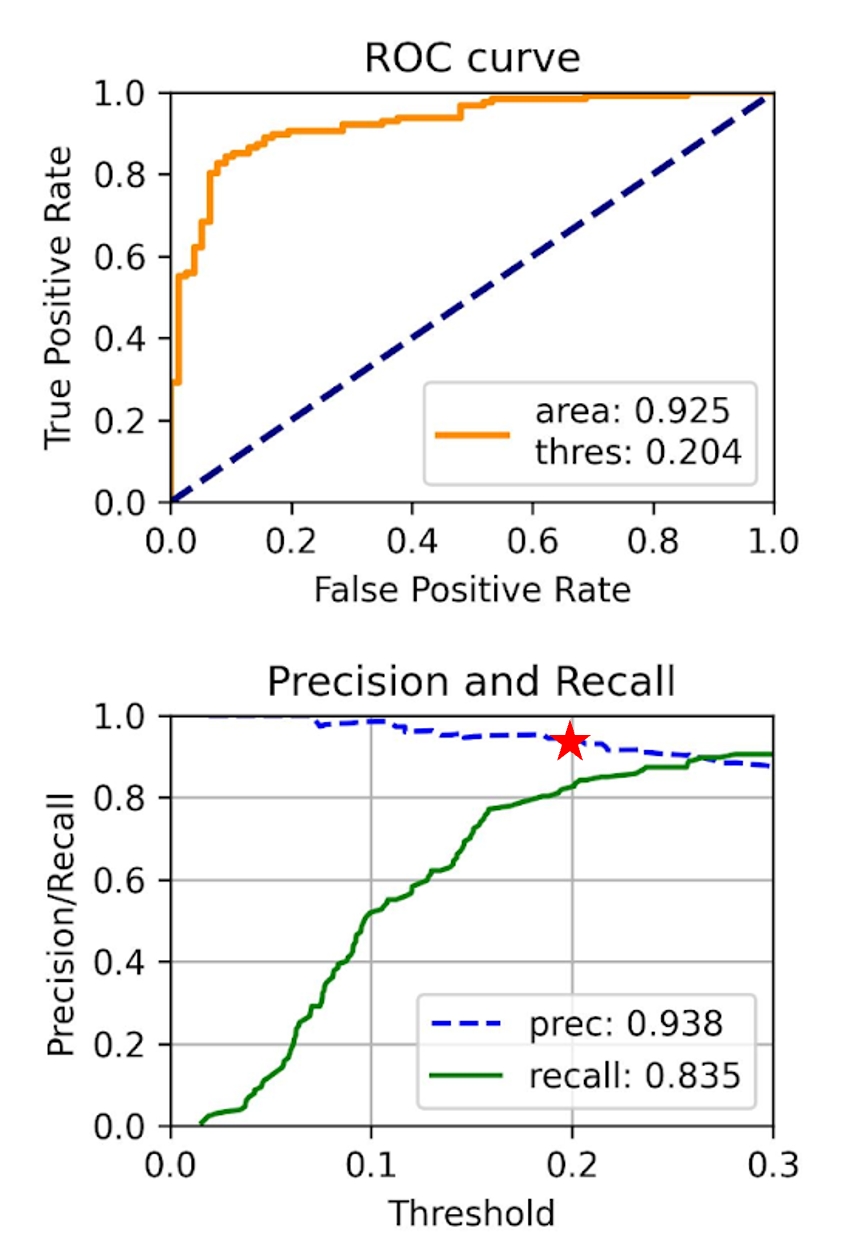}
    \caption{Threshold selection for identifying low-quality 3D reconstructions with unreasonable camera movements between consecutive frames. 
    \textbf{Left}: We label the correctness of {\raise.17ex\hbox{$\scriptstyle\sim$}}1k Colmap reconstructions via our interactive 3D annotation tool by reviewing the reconstruction result and the original video clip side-by-side. \textbf{Right}: We gather statistics (ROC curve, precision, and recall) on the distance of camera locations to the smoothed camera path from Kalman filter, and select a threshold (\textcolor{red}{red star}) that best separates correct and incorrect reconstructions.}
    \label{fig:annotate}
\end{figure}

\subsection{Data Filtering}
\label{subsec:data filter}

To filter out low quality data, we identify 3D reconstructions whose camera locations from neighboring frames are drastically apart (Fig.~\ref{fig:data} bottom), which is highly unlikely due to the continuous nature of camera movements. 
We consider the camera pose $\bm{c}$ and camera motion $\bm{a}$,
\begin{align}
    \bm{c} & = \left\{x, y, z, q_w, q_x, q_y, q_z\right\}, \\
    \bm{a} & = \left\{v_x, v_y, v_z, \omega_x, \omega_y, \omega_z\right\},
\end{align}
where $x, y, z$ denote the location and $q_w, q_x, q_y, q_z$ denote the rotation quaternion, both in Colmap convention. The velocity $v_x, v_y, v_z$ and the angular velocity $\omega_x, \omega_y, \omega_z$ are given by two consecutive frames. 

To estimate reasonable camera movements over neighboring frames, we first label the correctness of  {\raise.17ex\hbox{$\scriptstyle\sim$}}1k Colmap reconstructions. We then adopt Unscented Kalman Filter  \cite{wan2000unscented} to produce a smoothed camera path based on the original one, 
and compare the distance of original camera locations to the smoothed camera path to identify reconstruction with reasonable movements. We select a threshold that best separates the correct and incorrect reconstructions. 
We show our threshold selection process in Fig.~\ref{fig:annotate}.
Overall, the filtering process leaves us {\raise.17ex\hbox{$\scriptstyle\sim$}}99k samples for a total duration of {\raise.17ex\hbox{$\scriptstyle\sim$}}180 hours, 
a drastic drop from 1.5k hours of raw videos.

\begin{table}[t]
\resizebox{\linewidth}{!}{
\begin{tabular}{l|c|c|c|c|c}
\toprule
                & data source   & camera motion & accuracy & topic   & duration \\ \hline
Huang \etal \cite{huang2019learning2}     & online videos & optical flow     & low                & sports  & 0.6h     \\ 
Jiang \etal \cite{jiang2020example}     & 3D animations & human expert     & very high               & drama   & 0.4h     \\ 
DVCD18K \cite{ashtari2022drone}        & online videos & SLAM             & low                & general & 44.3h    \\ 
DroneMotion-99k & online videos & SfM + filtering  & high               & general & 182.3h   \\ \bottomrule
\end{tabular}
}
\caption{Comparison with existing Datasets. The DroneMotion-99k dataset uses real-world videos and an automatic method to extract camera trajectories and filter out low-quality data, which is inexpensive yet accurate. 
}
\label{tab:data}
\end{table}

We compare our final dataset with alternatives in Table~\ref{tab:data}. While departing from limited appearance changes in simulation data \cite{jiang2020example} with real-world data, DroneMotion-99k also contains high-quality 3D camera paths more accurate than those generated from optical flow \cite{huang2019learning2} or SLAM without camera intrinsics \cite{ashtari2022drone}.

\begin{figure*}
    \centering
    \includegraphics[width=\linewidth]{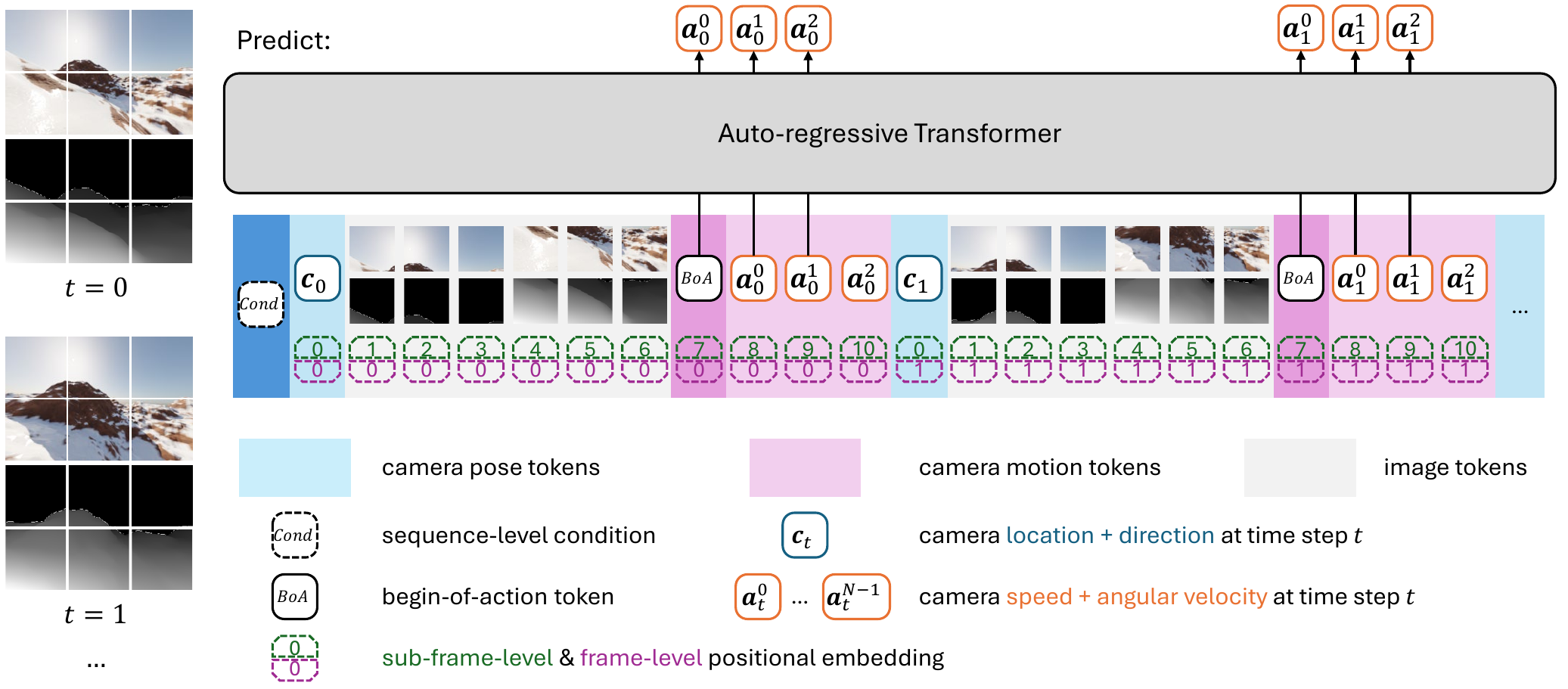}
    \vskip -0.05in
    \caption{Model overview of DVGFormer. To predict camera motion $\bm{a}_t$ for time step $t$, the auto-regressive architecture uses as input a long horizon with camera poses $\left\{\bm{c}_0, ..., \bm{c}_t \right\}$, motion $\left\{\bm{a}_0, ..., \bm{a}_{t-1}\right\}$, images $\left\{\bm{x}_0, ..., \bm{x}_t\right\}$ and their monocular depth estimations from all previous frames. Each action $\bm{a}_t$ is broken into $N$ intermediate steps $\left\{\bm{a}_t^0, ..., \bm{a}_t^{N-1}\right\}$ between time step $t$ and $t+1$. 
}
    \label{fig:model}
\end{figure*}

\section{The DVGFormer Model}
\label{sec:model}
We depict our model, DVGFormer, in Fig.~\ref{fig:model}. Built with an auto-regressive transformer, it considers all past frames, camera poses, and motions, and predicts camera motions for the next frame. 
Compared to generalist robotics models \cite{rt12022arxiv, rt22023arxiv}, this approach allows for a longer horizon and considers past camera poses and motions.  

\subsection{Input and Output}
In terms of input, first, we have a \texttt{<Cond>} token sampled from a random Gaussian noise as the overall condition for the entire video clip. 
For a time step $t$, we consider the camera pose $\bm{c}_t$ and the image $\bm{x}_t$ as state, and the camera motion $\bm{a}_t$ as action. 
Note that we break the action $\bm{a}_t$ into $N$ steps $\left\{\bm{a}_t^0, ..., \bm{a}_t^{N-1}\right\}$ at a $N$-times higher frame rate than the images to generate a smoother camera path while avoiding redundancy between consecutive frames. 
After tokenization, the combination of camera pose, motion, image patches, and a \texttt{<BoA>} (begin-of-action) token gives the overall representation for one frame. 
The \texttt{<BoA>} token marks that the next $N$ tokens will be used to predict the actions $\left\{\bm{a}_t^0, ..., \bm{a}_t^{N-1}\right\}$, which are then included as the next token for the auto-regressive transformer. 
Finally, tokens of past frames and the \texttt{<Cond>} token together forms the input to the auto-regressive transformer. 

In terms of output, the system predicts the 6 degree-of-freedom (6-DoF) camera motion $\bm{a}$ as a continuous value, which is normalized according to the statistics on the DroneMotion-99k dataset. 

During inference, the models predicts the next actions based on past information from the input. Then, we execute these actions to reveal the next camera pose $\bm{c}_{t+1}$ and image $\bm{x}_{t+1}$. We iterate this process to the generate future camera motions and get future frames. 
In practice, we update the image and camera pose at 3 fps, and predict $N=5$ camera motions for each image, making it essentially 15 fps. 
When paired with techniques like the recurrence mechanism and chunking \cite{dai2019transformerxl}, the auto-regressive transformer can generate camera trajectories with arbitrary length. 

\subsection{Tokenization}
\label{subsec:token}
For camera poses and motions, we use Multi-Layer Perceptron (MLP) for their tokenization. 
We do not discretize them  \cite{rt12022arxiv,rt22023arxiv} since it can drastically increase the sequence length when the video unfolds. 

For image patch tokens, we use a two branch design. First, we use DINOv2 \cite{oquab2023dinov2} to extract features from RGB image patches and train MLP layers to project them. Secondly, to inject 3D information to the model, we consider the monocular depth estimation results from Depth-Anything \cite{depth_anything_v2}  and train convolutional layers to tokenize the predicted depth map. For each patch, the feature vectors from the two branches are then added together. We take a resolution of $168\times 298$ for the images and downsample the final feature map  to a resolution of $5\times 9$. 

For positional embeddings,  sequential orders  at both  frame level and sub-frame level are very important  for informing the time step $t$ and the relative position of  tokens within each time step. Therefore, we introduce a bi-level design with frame-level and sub-frame-level  positional embeddings, which are interleaved to formulate the overall positional embedding. 
Compared to using   different tokens for different locations in the sequence \cite{janner2021sequence,rt12022arxiv,rt22023arxiv} , this bi-level design introduces less parameters and is  less susceptible to overfitting. Compared to the frame-level-only design \cite{chen2021decision}, this approach can differentiates different tokens within one time step, making it possible to contrast the same image patch over time for motion extraction.  
We choose a maximum duration of 10 seconds in DVGFormer, resulting in 30 different frame-level positional embeddings with the image updated at 3 fps, and 52 different sub-frame-level positional embeddings with 1 camera pose token, 45 image tokens, 1 \texttt{<BoA>} token, and 5 action tokens per frame. 

\subsection{Loss Functions}
\label{subsec:loss}

We use the standard sequence-level loss with casual masking to train DVGFormer. We apply $\mathcal{L}_1$ loss to supervise the continuous prediction of the actions. 


\section{Experiments}
\label{sec:experiments}

\begin{figure*}
    \centering
    \includegraphics[width=\linewidth]{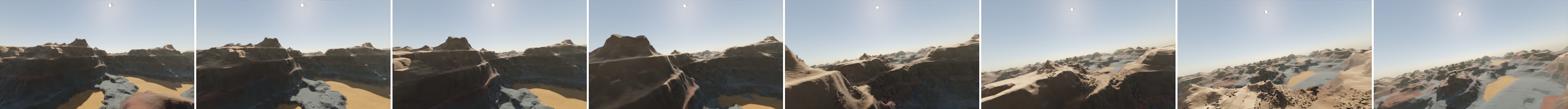}
    \includegraphics[width=\linewidth]{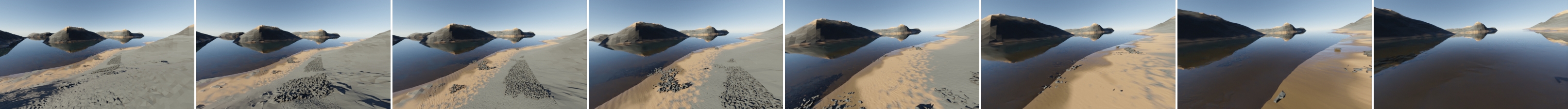}
    \includegraphics[width=\linewidth]{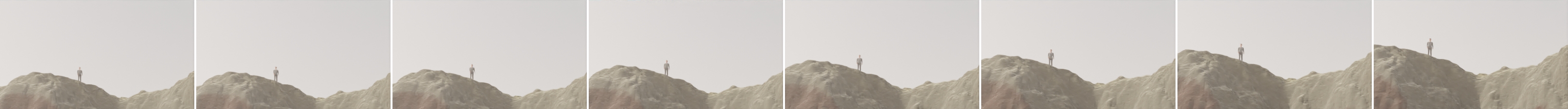}
    \includegraphics[width=\linewidth]{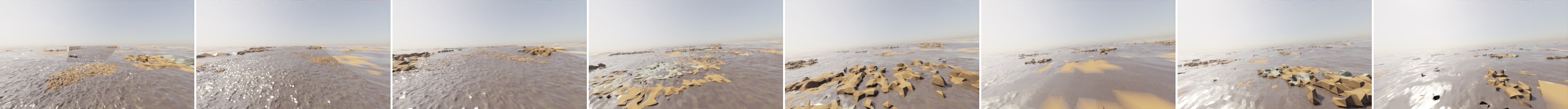}
    \includegraphics[width=\linewidth]{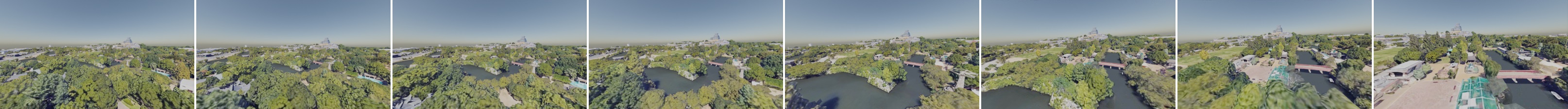}
    \includegraphics[width=\linewidth]{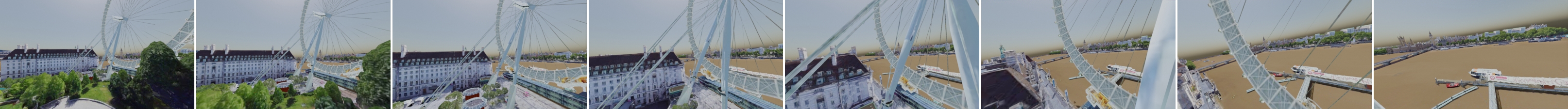}
    \includegraphics[width=\linewidth]{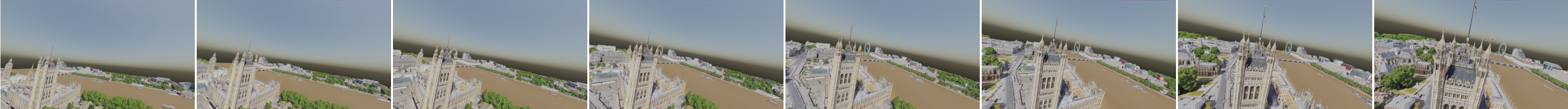}
    \includegraphics[width=\linewidth]{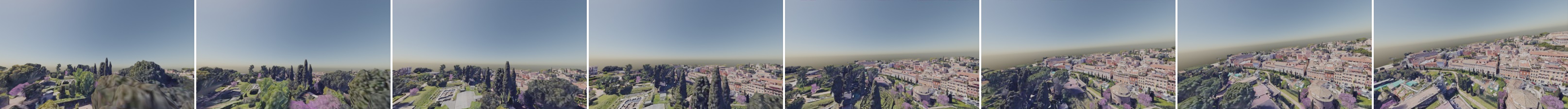}
    \includegraphics[width=\linewidth]{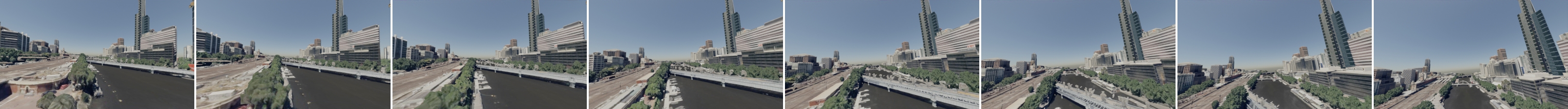}
    \includegraphics[width=\linewidth]{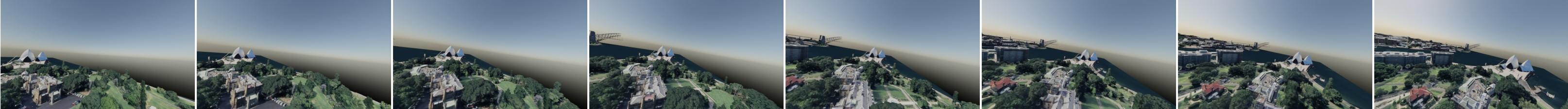}
    \caption{Visualization of the recorded videos. DVGFormer learns techniques like keeping the actor in frame, navigating through obstacles, maintaining low altitude to increase perceived speed, orbiting tower and buildings, or increasing altitude and pitching down camera for a full view, all directly from the DroneMotion-99k dataset and without any heuristics. }
    \label{fig:qualitative}
\end{figure*}

\subsection{Implementation Details}
\label{subsec:training}
We represent  camera poses in global coordinates specified by the first frame, and camera motions in the local coordinates specified by the current camera pose. 
We use the GPT-2 architecture \cite{radford2019language} for the backbone and randomly initialize 12 layers and 6 heads at 384 hidden dimensions, marking a total of 40M parameters.  
We use the ViT-Small \cite{dosovitskiy2021imageworth16x16words} version of both  DINOv2 \cite{oquab2023dinov2} and Depth-Anything \cite{depth_anything_v2}. 

We only train the MLP tokenizers for the camera pose and motion, the MLP projection layers for DINOv2 features, the convolutional layers for monocular depth estimation results, the bi-level positional embeddings, and the transformer itself. 
We apply random horizontal flip as data augmentation, which flips not only the 2D images but also the 3D camera trajectories (pose and motion). 
Experiments are run on two NVIDIA RTX-3090 GPUs with a batch size of 32 per GPU and 4 gradient accumulation steps. 

\subsection{Evaluation Platform}
\label{subsec:eval platform}
To evaluate the predicted camera trajectories for drone videography, we first build an interactive evaluation platform. Specifically, we use Blender \cite{blender}, an open-source software for 3D content creation, which can render images of 3D assets based on the camera configurations. The simulation environment takes the camera motion as input, and returns the next camera pose and the rendered image. 
It also detects collision with the 3D scene and use it as a signal for terminating the video. 
We use existing 3D assets  on natural and civic scenes for evaluation. 
We adopt InfiniGen \cite{infinigen2023infinite} for generating natural scenes randomly, and use BLOSM toolkit \cite{blosm} to import Google Earth \cite{GoogleEarth} 3D scans of cities. 
Overall, we collected 38 random natural scenes and 7 cities across Asia, Europe, America, and Oceania.

By default, we generate 10 second duration videos with DVGFormer. We also include 20 seconds generation results to demonstrate the capability of arbitrary length generation with chunking \cite{dai2019transformerxl}.

\subsection{Baseline}
\label{subsec:baseline}
We compare against an RT-1 \cite{rt12022arxiv} inspired architecture that considers the past 6 frames at 3 fps. Note that for a fair comparison, we use the same DINOv2 image feature in contrast to the EfficientNet feature with language injection in the original RT-1. Compared to our approach, the main differences lie in \textbf{a)} the limited temporal receptive field, 2 seconds in RT-1 vs up to 10 seconds in DVGFormer, and \textbf{b)} the tokenization, where RT-1 only includes image tokens and DVGFormer additionally considers both camera pose and motion. These differences align with our design highlights in DVGFormer.

\begin{table}[]
\centering
\resizebox{\linewidth}{!}{
\begin{tabular}{l|c|c|c|c}
\toprule
           & preference$\uparrow$ & collision$\downarrow$ & $\Delta v \downarrow$ & $\Delta \omega \downarrow$ \\
           \hline
RT-1 inspired      & 29.5\%                &  33.7\%         & 1.2\%  & 18.4\%      \\
DVGFormer &   70.5\%              & 15.2\%          &  0.7\%  & 16.8\%   \\
\bottomrule
\end{tabular}
}
    \caption{Comparison with the RT-1 \cite{rt12022arxiv} inspired baseline. Videos recorded with DVGFormer have higher user preference, lower collision rate and smoother camera motions.}
    \label{tab:comparison}
\end{table}

\begin{figure*}
    \centering
    \includegraphics[width=0.2\linewidth]{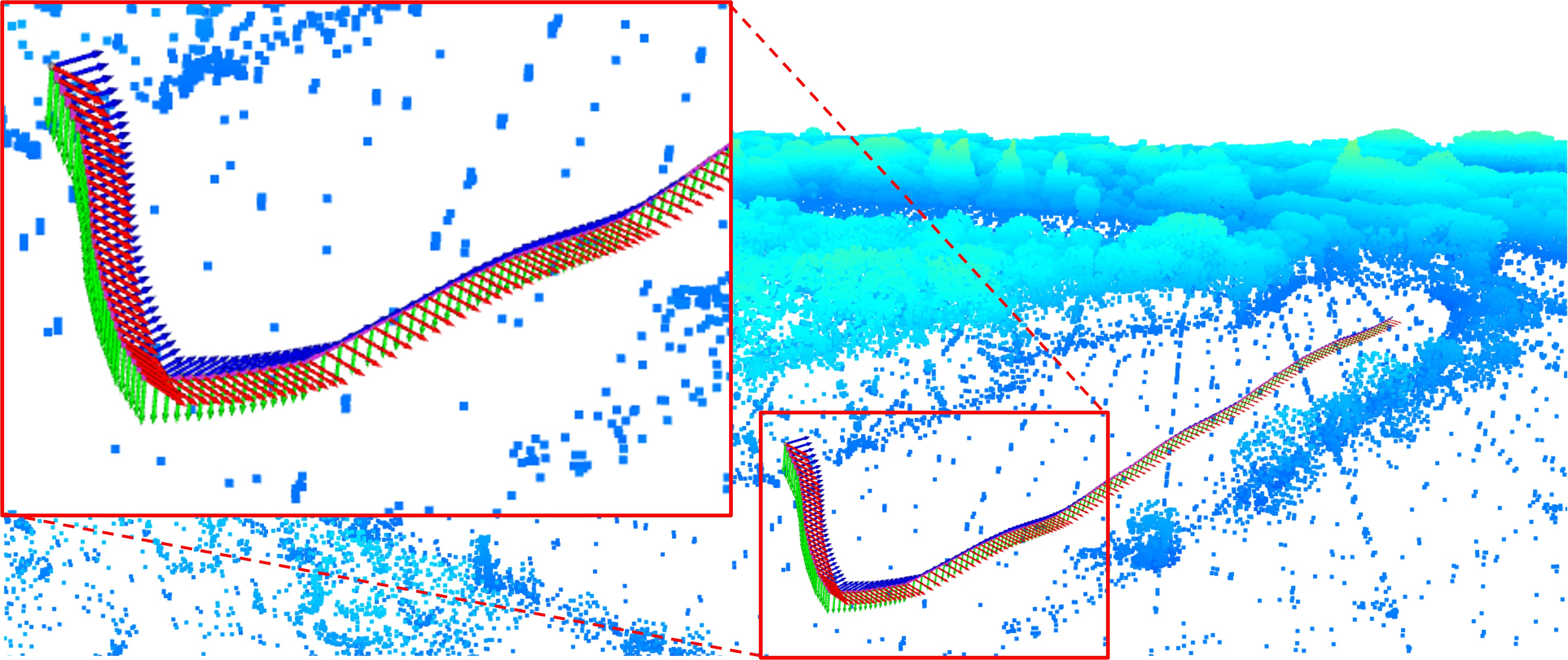}
    \hfill
    \includegraphics[width=0.76\linewidth]{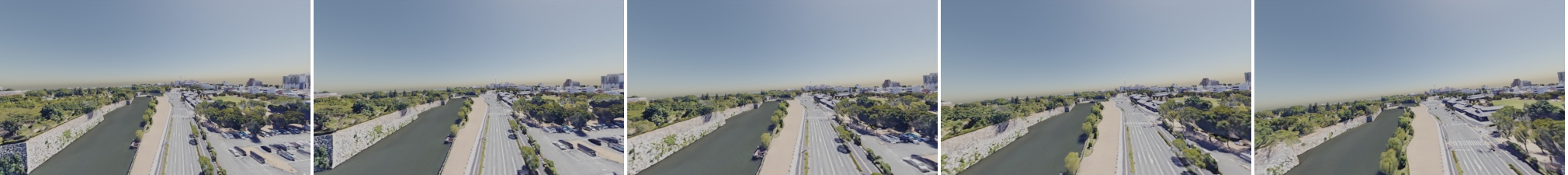}
    \includegraphics[width=0.2\linewidth]{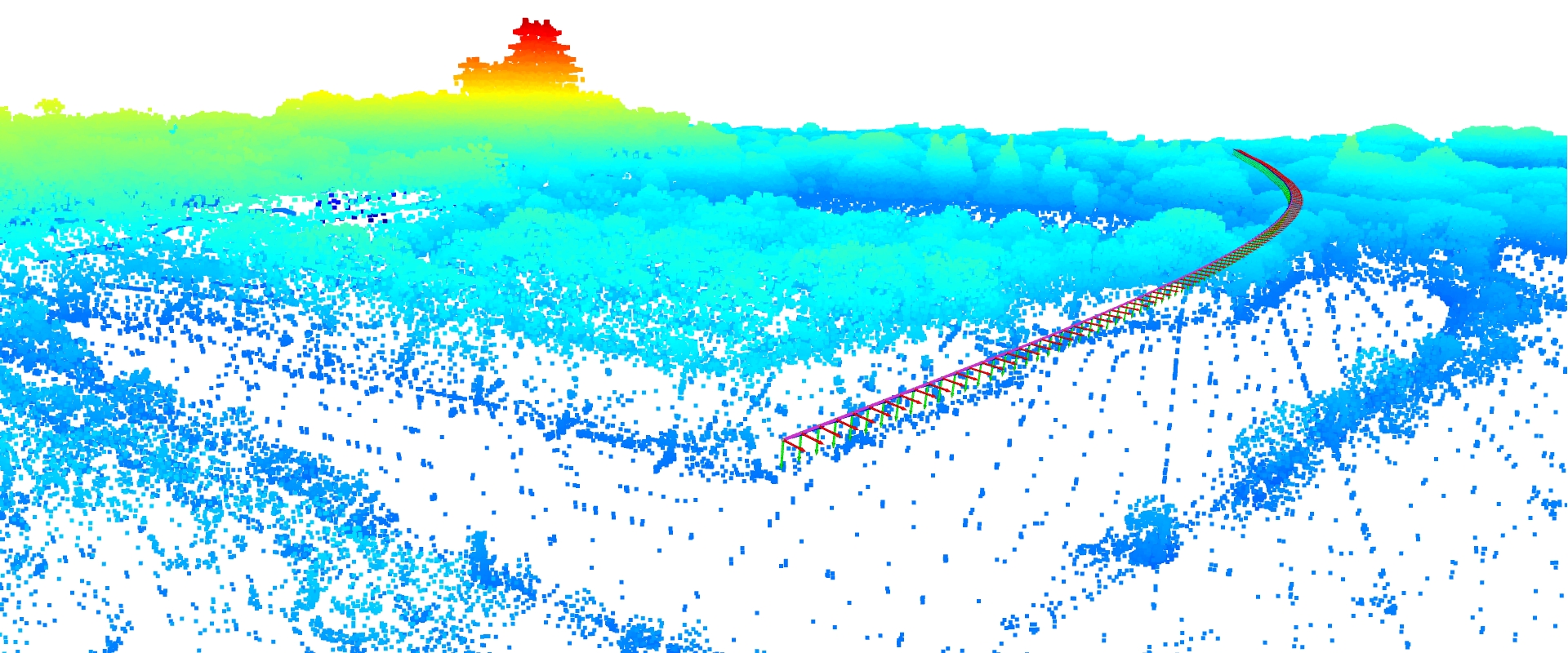}
    \hfill
    \includegraphics[width=0.76\linewidth]{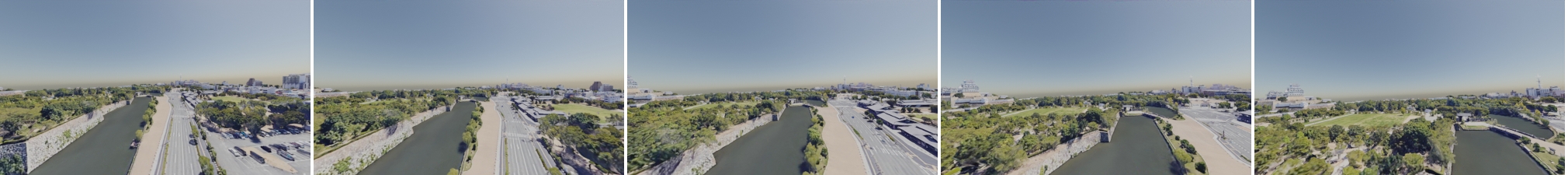}
    \vskip -0.05in
    \caption{3D camera trajectory comparison. Compared to the RT-1 inspired basline (\textbf{top}), using a long horizon and the previous camera path helps DVGFormer (\textbf{bottom}) produce a smoother camera path without sudden movements or direction changes (\textcolor{red}{red box}). 
    }
    \label{fig:compare3d}
\end{figure*}

\begin{figure*}
    \centering
    \includegraphics[width=0.2\linewidth]{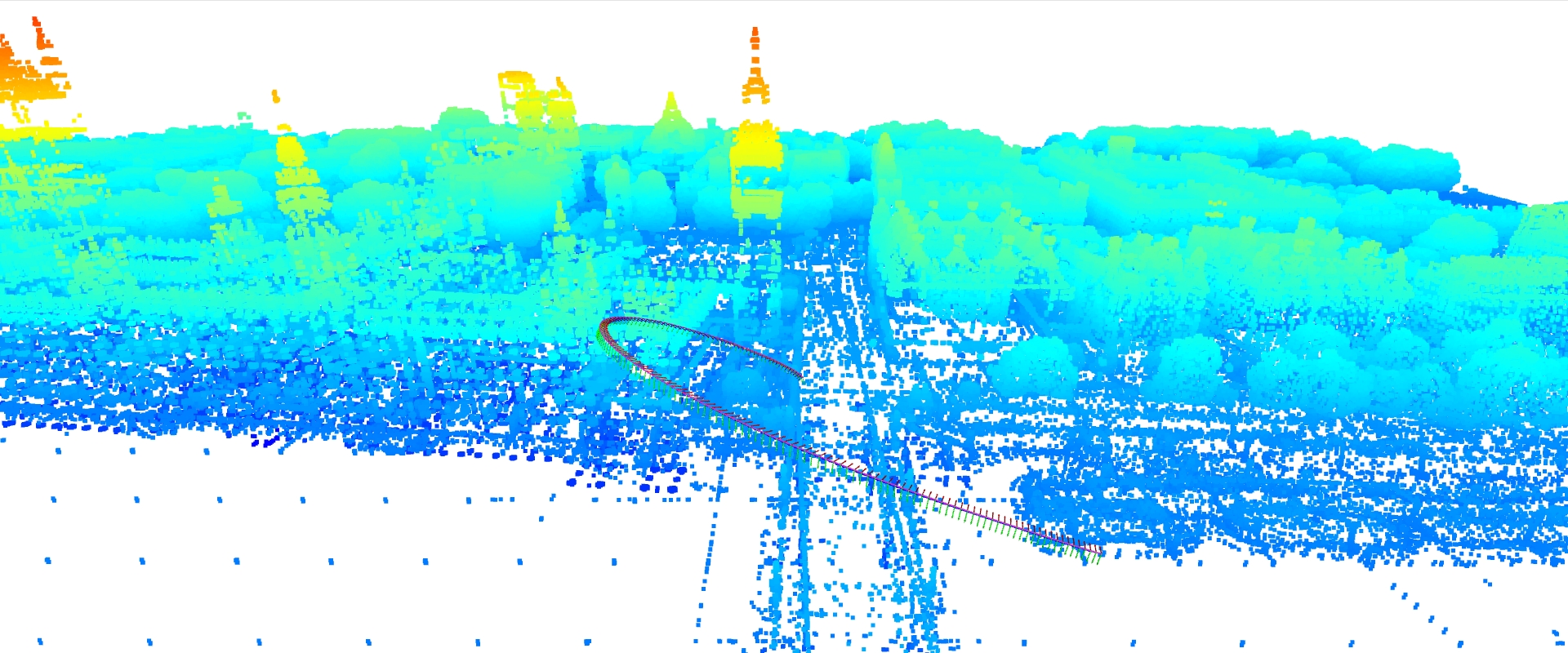}
    \hfill
    \includegraphics[width=0.2\linewidth]{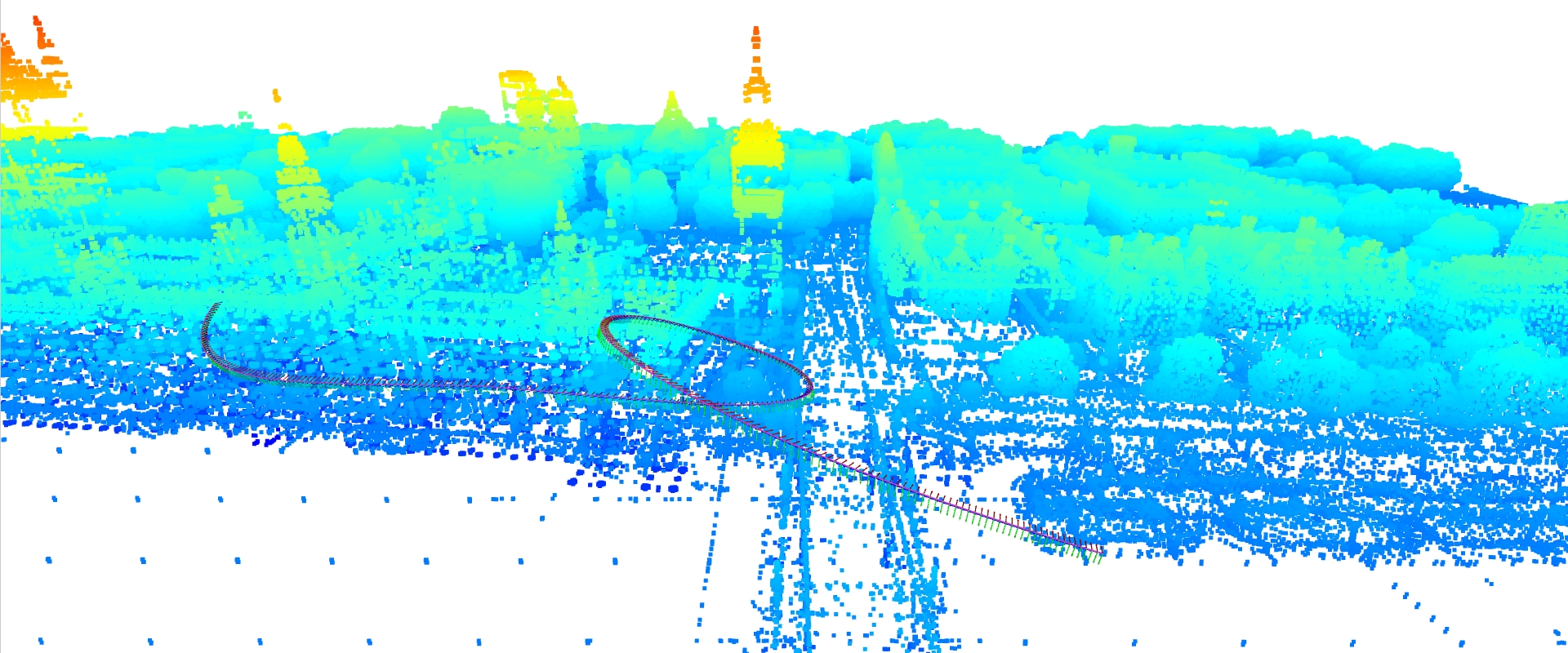}
    \hfill
    \includegraphics[width=0.58\linewidth]{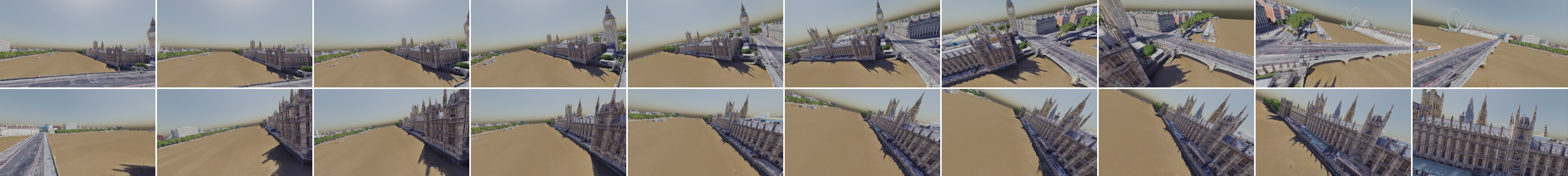}
    \caption{Generating 10-second and 20-second videos. By default, DVGFormer has a maximum duration of 10 seconds (\textbf{left}). However, using the recurrence mechanism \cite{dai2019transformerxl}, it can continue camera path generation (\textbf{middle}) and produce 20-second videos (\textbf{right}). 
    }
    \label{fig:20s}
\end{figure*}

We skip the previous heuristic-based methods \cite{10.1145/237170.237259,azzarelli2024reviewing,joubert2016towards, nageli2017real, huang2018act} as the heuristics requires human actors and cannot generalize to many test scenarios. 

\subsection{Quantitative Study}
\label{subsec:quant}

In this section, we evaluate the generated camera trajectory according to their user preference, collision rate, and the trajectory smoothness. Specifically, we measure the trajectory smoothness in terms of the maximum relative change in velocity $\Delta v$ and angular velocity $\Delta \omega$, since these sudden changes can greatly affect the subjective feelings. 

In Table~\ref{tab:comparison}, we find that the proposed method offers much higher user preference compared to the baseline. 
It has a lower crash rate, demonstrating its effectiveness in 3D awareness. This partially contributes to the user preference since most would favor videos without collision. 
The camera trajectories generated by the proposed method also are smoother. Without specifically considering the camera poses and motions in previous frames, the RT-1 inspired baseline cannot guarantee the motion consistency over neighboring frames. One other issue is the limited horizon in the baseline, because the 6 past frames over 2 seconds cannot provide enough information over the entire video. This issue is less pronounced in other robotics tasks like navigation \cite{shah2023vint} or picking, moving, and placing objects \cite{rt12022arxiv,rt22023arxiv}, because their focus would be on the successful execution of a certain task, not the motion consistency across frames. However, for videography, a smooth camera trajectory could greatly affects the user perception, thus separating beginners and professionals. 
These quantitative results verify the effectiveness of our design.

\begin{figure*}
    \centering
    \includegraphics[width=\linewidth]{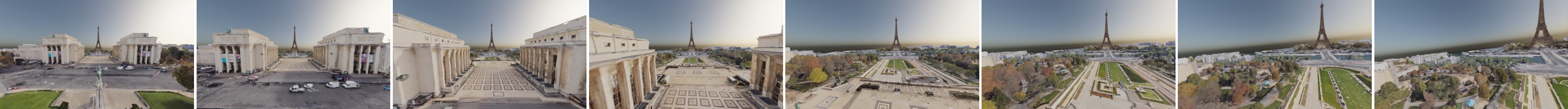}
    \vspace{1mm}
    \includegraphics[width=\linewidth]{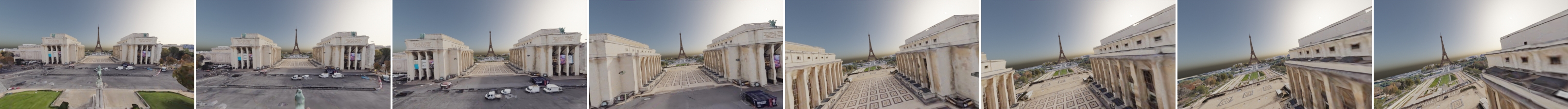}
    \includegraphics[width=\linewidth]{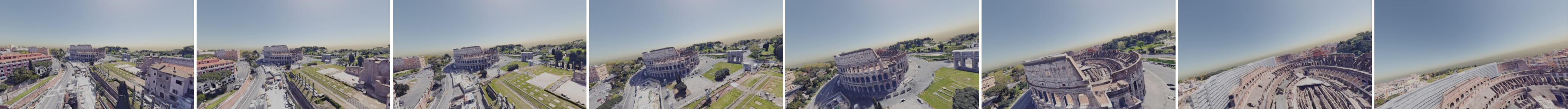}
    \includegraphics[width=\linewidth]{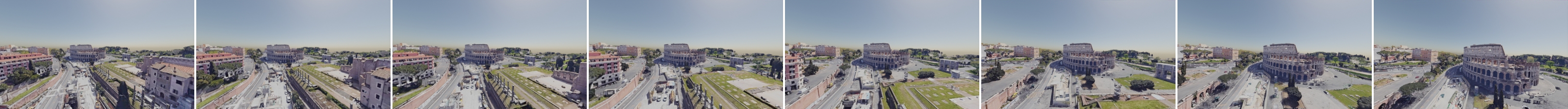}
    \vskip -0.05in
    \caption{
    From the same initial image (leftmost), our model can output different but feasible camera paths using different \texttt{<Cond>} tokens. Two top rows and  bottom rows shows show two different examples. }
    \label{fig:controllable generation}
\end{figure*}

\begin{figure*}
    \centering
    \includegraphics[width=\linewidth]{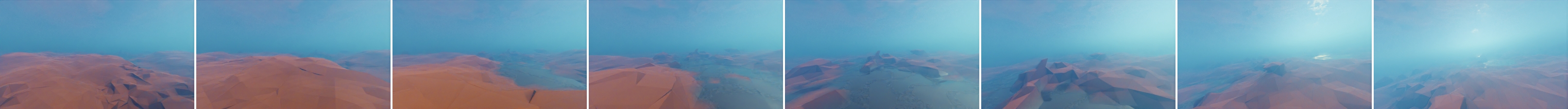}
    \vskip -0.05in
        \caption{Generalization to unseen scenarios. Trained from drone videos only, the proposed model can also work in underwater scenes. }
    \label{fig:underwater}
\end{figure*}

\subsection{Qualitative Study}
\label{subsec:quality}
We demonstrate video snippets generated by the proposed method in Fig.~\ref{fig:banner} and Fig.~\ref{fig:qualitative}. For natural scenes, we find the AI cameraman learns to adjust the camera path according to the terrain and avoid obstacles. It also demonstrated several other techniques. For example, it learns to  maintain a low altitude for flyovers and keeping close to the ground to increase the perceive speed and the stimulating effect from the motions. Even without any specifically designed heuristics, it learns to keep to the actor in frame while panning the camera for reveal. For city scenes, the AI drone cameraman also successfully navigates around the buildings most of the time and can adjust the motion patterns based on the scenes, like adjusting the trajectory to fly through the London Eye or to follow the Yarra River  in Fig.~\ref{fig:qualitative}. Some of the other worth-mentioning techniques from DVGFormer includes orbiting moves while approaching architectures like the Sydney Opera House in Fig.~\ref{fig:banner} and the tower at the Palace of Westminster in Fig.~\ref{fig:qualitative}.

We directly compare the 3D camera trajectory from the RT-1 inspired baseline and the proposed method in Fig.~\ref{fig:compare3d}. The baseline struggles to generate a smooth 3D path, leading to stuttering and sudden movements in the rendered videos. This is also reflected by the more drastic changes in velocities and angular velocities in Table~\ref{tab:comparison}. In comparison, DVGFormer can predict a smooth camera curve for revealing the Himeji Castle.

We show an example of predicted camera paths for different video durations in Fig.~\ref{fig:20s}. When given a longer duration, the proposed model can extend the existing trajectory in a very smooth manner. 

We can also control the generated camera path via the \texttt{<Cond>} token, as shown in Fig.~\ref{fig:controllable generation}. When feeding different random noise as the \texttt{<Cond>} token for the same starting frame, the generated camera trajectories are also affected. 

In terms of generalization ability, when applied to an underwater scenario, the proposed DVGFormer can still translate its learned experience in drone videography to produce similar videos under water (Fig.~\ref{fig:underwater}). 

\section{Discussions and Limitatinons}
\label{sec:discuss}

\textbf{Contribution statement.}
In this study, we build a scalable approach to learning the camera movement control for drone videography by introducing a real-world dataset and a model architecture. In terms of \textit{data}, DroneMotion-99k escapes the limited appearance changes in simulation training \cite{shah2018airsim,jeon2020detection,yu2022enabling} and introduce an automatic process to retrieval ground-truth 3D camera paths without having to record teleoperations from human experts \cite{rt12022arxiv,walke2023bridgedata}. 
In terms of \textit{model}, DVGFormer considers a longer horizon and richer information (previous camera path) when compared to current generalist robotics models \cite{rt12022arxiv,rt22023arxiv}. 
Compared to previous studies that also use auto-regressive transformer to consider all past frames \cite{chen2021decision,janner2021sequence}, we tackle an arguably more difficult problem than balancing a walker robot or playing simple games \cite{openaigym} with even fewer feedback (no clear rewards). Compared to architectures for video understanding \cite{carreira2017quo,gberta_2021_ICML,wang2021self}, instead of sampling a fixed number of frames regardless of the video duration, the network is updated with image and previous camera path at a fixed frame rate to avoid confusing how fast the scene changes.

\textbf{Scale ambiguity of 3D reconstruction} is a known issue, and thus we use the drone speed for scale normalization. We also experimented with metric depth estimation \cite{yin2023metric3d} to solve the scale ambiguity, but it fails to produce reliable results and needs further investigation.

\textbf{Choice of 3D perception.} We currently use monocular depth estimation for 3D perception, which only operates on a \textit{per-frame} basis. Although they are less usable for dataset creation due to the absent camera intrinsics, SLAM methods could be very effective in inference-time since they can provide \textit{sequence-level} 3D information and the camera intrinsics are available during testing.

\textbf{Other sensors} like GPS, Inertial Measurement Unit (IMU), LiDAR, or proximity sensors  are not considered in this study due to their unavailability from online videos. With that said, we believe they can be further included to in future works to further improve the system. 

\textbf{Dependency on previous motions} is a side effect of the auto-regressive architecture, since it can cause the model to prioritize the camera trajectory smoothness over collision avoidance. The behavior cloning approach also partially contributes to this because there are only success cases and no failure case to warn the model against crashing. 

\textbf{Implementation on drone hardware.} For piloting a drone in the real world, the current approach still has an undesirable crash rate. We expect future work to combine collision avoidance into the overall optimization target. 


\section{Conclusion}
\label{sec:conclusion}
In this paper, we focus on recording \textit{existing} subjects into attractive videos via AI videography. We study camera movement controls and introduce an scalable approach with real-world data and a generalizable model. The former includes 99k high-quality camera trajectories for a total duration of 180 hours, and an automatic pipeline to reconstruct and filter the 3D camera path. The latter is built on latest vision foundation models and the auto-regressive transformer architecture, and jointly considers path camera path and images over a long horizon. During simulation testing, the proposed approach successfully generates camera trajectories for capturing aesthetically pleasing videos.


\section*{Acknowledgment}
This research was supported by the ARC Discovery Project (DP210102801). We thank Google Cloud for the gift funds.


\appendix
\section*{Appendix}

\begin{figure}[h]
    \centering
    \includegraphics[width=\linewidth]{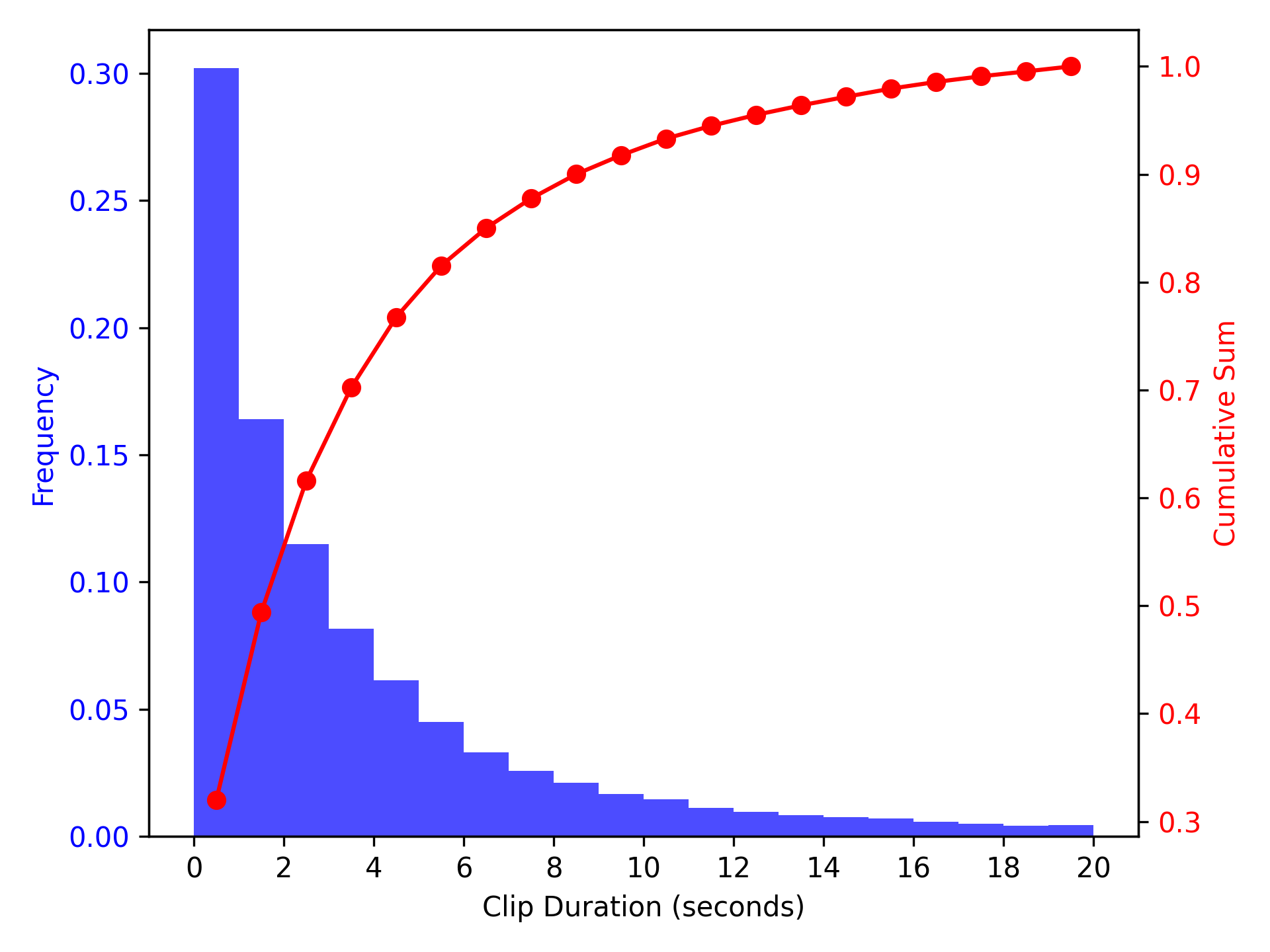}
    \caption{Clip duration distribution. 86.6\% of the clips produced by shot change detection algorithm \cite{PySceneDetect} are 10 seconds or shorter. 69.4\% of the clips are 1 second or longer. }
    \label{fig:clip duration}
\end{figure}

\section{Data Collection Details}

\textbf{Scrapping videos.}
We collect the dataset from YouTube videos. We search videos with key words including ``cinematic drone videos'', ``cinematic drone footage'', ``cinematic drone footage 4k'', ``cinematic fpv footage 4k'', \etc, where FPV stands for first-person-view, a specific shot types that feature drastic perspective changes to provide stimulating visual effects.

After searching YouTube videos, we first filter out ones with sensitive information including weaponized or harmful usage, for which we consider a blacklist of words including ``weapon'', ``soldier'', ``attack'', ``strike'', ``military'', ``surveillance'', ``horror'', \etc. 

Next, we download the videos with the yt-dlp \cite{ytdlp} package. We collected a total of 13,653 videos for a total duration of 1,485 hours.

\textbf{Video data filtering and clip generation.}
We only preserve those in the landscape mode and drop the portrait mode videos, since the natural sensor arrangement is the landscape mode. While 2,820 out of 13,653 videos are in the portrait mode, their total duration is much shorter in comparison, only 19.9 hours, which constitutes of 1.3\% of the total duration of 1,485 hours. This makes the following procedures much easier because the remaining videos are all of similar aspect ratios. 

\begin{table}[]
\resizebox{\linewidth}{!}{
\begin{tabular}{c|c|c|c}
\toprule
raw videos & w/o portrait & clips w/o dialog & after filtering \\ \hline
1,485      & 1,465        & 867              & 182                     \\ \bottomrule
\end{tabular}
}
\caption{Total video duration (hours) of the collected dataset at different stage of the processing pipeline.}
\label{tab:duration}
\end{table}

We then run PySceneDetect \cite{PySceneDetect} to detect the shot changes in videos, which produces a total of 642,806 individual clips. We show their duration distribution in Fig.~\ref{fig:clip duration}. We filter out 30.6\% of all clips whose duration is shorter than 1 second since they are too short and can be very difficult for the reconstruction. These sub-one-second fragments only make up for 2.1\% of the total video duration. For the 13.4\% of all clips whose duration is longer than 10 seconds, we break them down into partial clips with a maximum duration of 10 seconds. 

We filter out clips with dialog using the automatically generated or uploaded closed captions in YouTube videos. Apart from a few words in a whitelist, \eg, ``[music]'', ``[silence]'', ``[background noise]'', ``[pause]'', ``[sound effect]'', the closed caption often suggest that the scene is unrelated with drone videos. After this step, there are 449,997 clips remaining for a total duration of 867 hours. 

\textbf{3D reconstruction.}
We extract frames from videos at a resolution of 1080p and a frame rate of 15 fps. In Colmap \cite{schoenberger2016sfm}, we consider 512 feature points and extract SIFT \cite{lowe2004distinctive} features with \texttt{estimate\_affine\_shape=True} and \texttt{domain\_size\_pooling=True}. We also ignore the focal length changes in videos and set \texttt{single\_camera=True} since we only focus on the change of camera location and direction in this work. We use \texttt{guided\_matching=True} and the \texttt{sequential\_matcher} for the frames extracted from videos. We run up to two sparse reconstruction steps, and when the first sparse reconstruction does not return any result, we extend some of the requirements and run a second round. This produces 264,596 reconstructions. 

\begin{table*}[]
\centering
\begin{tabular}{l|l}
\toprule
function                    & specification                                         \\ \hline
camera pose tokenization    & 3 MLP layers, 384 dimensions                                  \\ 
camera motion tokenization  & 3 MLP layers, 384 dimensions                                  \\ 
RGB feature projection      & AvgPool(12, 21), 2 MLP layers, 384 dimensions                 \\ 
depth feature               & AvgPool(12, 21), 3 convolutional layers, 3x3 kernel, 384 dimensions    \\ 
\texttt{<BoA>} token                   & $1\times384$ vector \\ 
frame-level PE              & $30\times192$ matrix (10 seconds at 3 fps for images)                                              \\ 
sub-frame-level PE          & $52\times192$ matrix (1 pose token, 45 image tokens, 1 \texttt{<BoA>} token, 5 motion tokens)                                              \\ 
auto-regressive transformer & GPT-2 architecture, 12 layers, 6 heads, 384 dimensions \\ \bottomrule
\end{tabular}
\caption{Details of the learnable modules in DVGFormer.}
\label{tab:parameter}
\end{table*}

\textbf{Reconstruction filtering.}
Reconstructions with duration smaller than 15 frames or 1 second are discarded. We normalize the data according to the average difference in locations (speed). We discard data whose maximum camera speed exceeds 3 times the average camera speed. 

We consider the reconstructions whose camera locations from neighboring frames are drastically apart as low-quality data. We design an automatic process for identifying and discarding those low-quality reconstructions. Specifically, we use Kalman filter to estimate a smoothed version of the camera trajectory, which then discloses how different it is compared to the original camera trajectory. If these two trajectories are drastically different, we then automatically discard the 3D reconstruction. 

During implementation, we choose Unscented Kalman Filter (UKF) \cite{wan2000unscented} in FilterPy \cite{filterpy}. We consider a  representation with 13 dimensions by combining the 7-dimensional translation vector and rotation quaternion in camera pose $\bm{c}$ and the 6-dimensional speed and angular speed in camera motion $\bm{a}$. We set $\alpha=0.1$, $\beta=2$, and $\kappa=10$ for the hyperparameters in UKF. 

Based on 1k labeled data from our interactive labeling tool, we select the optimal threshold for the difference between the original and the smoothed camera trajectory at 0.2 (see Fig. 3 in the main paper). We end up with 99,003 camera trajectories with a total duration of 182 hours. 

We compare the total video duration at each stage of the data processing pipeline in Table~\ref{tab:duration}.

\section{Model Details}
For camera pose and motion tokenization, we adopt 3 MLP layers with 384 hidden dimensions. For DinoV2 \cite{oquab2023dinov2} feature projection, given image resolution of $168\times 294$,  we first downsample the feature map from $12 \times 21$ to $5 \times 9$ with average pooling. Then, we apply two MLP layers with 384 hidden dimensions. For the monocular depth estimation results from DepthAnything \cite{depth_anything_v2}, we also average pool the depth map from $168\times 294$ to $5 \times 9$, and apply three convolutional layers with $3\times 3$ kernel size and 384 hidden dimensions. We use GELU activation \cite{hendrycks2016gaussian} for all modules unless specified. For positional embeddings (PE), we consider 192 dimensions for both frame-level PE and sub-frame-level PE, and concatenate them together into an overall PE of 384 dimension before adding to the tokens. 
We list all learnable modules in DVGFormer in Table~\ref{tab:parameter}.

\section{Experiment Details}
For synthetic natural scene in simulation experiments, we use InfiniGen \cite{infinigen2023infinite} to randomly generate 38 scenes from 10 pre-defined natural scene types \texttt{arctic, canyon, cliff, coast, desert, forest, mountain, plain, river, snowy\_mountain}. We use the \texttt{simple},  \texttt{no\_assets}, and \texttt{no\_creatures} settings to ensure the generated scene is not too complex. For generalization ability study in Fig. 9 in the original paper, we the \texttt{under\_water} setting in InfiniGen. For scenes with human actor, we import free assets downloaded from the SketchFab website \cite{sketchfab}. 

For real cities in simulation experiments, we choose London, Paris, Rome, New York, Sydney, Melbourne, and Himeji. Regions of roughly 1km $\times$ 1km area are manually selected and the corresponding Google Earth 3D meshes are imported via the BLender Open StreetMap (BLOSM) toolkit \cite{blosm}. 

During inference, we render the scenes with $225\times 400$ resolution and 64 samples with a camera sensor width of 36mm. The lower resolution and number of samples in Blender \cite{blender} helps to increase the rendering speed. With that said, we can select an arbitrarily high resolution with higher sample quality during the final rendering, since we do not modify any 3D assets or 2D pixels and the videos are faithful depiction of the \textit{existing} scene.

We do not use the image generation metrics like PSNR or text-to-video evaluation metrics in VBench\cite{huang2024vbench}. In terms of 2D image, since we do not modify 2D pixel, the 2D image PSNR would only refect the rendering quality in Blender, which itself is adjustable based on the computational budget. Same goes for the image-based metrics of  Appearance Style, Scene, Color, Multiple Objects, Object Class, and Imaging Quality in VBench. As for temporal evaluation for videos, VBench metrics including Motion Smoothness, Temporal Flickering, Background Consistency, Subject Consistency, Overall Consistency, are not applicable either, since the the temporal consistency between neighboring frames are also guaranteed by the video rendering pipeline in Blender.

We report the user preference and collision rate in Table 2 in the main document on 184 videos. These 184 videos cover 38 natural scenes from InfiniGen and 7 real city scans. On both the DVGFormer and the baseline method, we use the same initial camera pose for each video. 

{
    \small
    \bibliographystyle{ieeenat_fullname}
    \bibliography{main}
}


\end{document}